\documentclass{article}

\usepackage[preprint]{neurips_2019}

\usepackage[utf8]{inputenc} % allow utf-8 input
\usepackage[T1]{fontenc}    % use 8-bit T1 fonts
\usepackage{hyperref}       % hyperlinks
\usepackage{url}            % simple URL typesetting
\usepackage{booktabs}       % professional-quality tables
\usepackage{amsfonts}       % blackboard math symbols
\usepackage{nicefrac}       % compact symbols for 1/2, etc.
\usepackage{microtype}      % microtypography
\usepackage{threeparttable}
\usepackage{amsmath,amsfonts,amsthm} % Math packages
\usepackage{multirow}

\usepackage{algorithm}
\usepackage{algorithmic}
\usepackage{amssymb}
\usepackage{graphicx}

\title{A Meta Approach to Defend Noisy Labels by the Manifold Regularizer PSDR}

\author{%
	Pengfei Chen$^1$,~
	Benben Liao$^2$,~
	Guangyong Chen$^2$\thanks{Correspondence to: Guangyong Chen <gycchen@tencent.com>.},~
	\textbf{Shengyu Zhang}$^{2}$ \\
	The Chinese University of Hong Kong$^1$,~
	Tencent$^2$\\
}

\begin{document}

\maketitle

\begin{abstract}
	Noisy labels are ubiquitous in real-world datasets, which poses a challenge for robustly training deep neural networks (DNNs) since DNNs can easily overfit to the noisy labels. Most recent efforts have been devoted to defending noisy labels by discarding noisy samples from the training set or assigning weights to training samples, where the weight associated with a noisy sample is expected to be small.  Thereby, these previous efforts result in a waste of samples, especially those assigned with small weights. The input $x$ is always useful regardless of whether its observed label $y$ is clean. To make full use of all samples, we introduce a manifold regularizer, named as Paired Softmax Divergence Regularization (PSDR), to penalize the Kullback-Leibler (KL) divergence between softmax outputs of similar inputs. In particular, similar inputs can be effectively generated by data augmentation. PSDR can be easily implemented on any type of DNNs to improve the robustness against noisy labels. As empirically demonstrated on benchmark datasets, our PSDR impressively improve state-of-the-art results by a significant margin.
\end{abstract}

\section{Introduction}
\label{Sec_Introduction}
DNNs have gained a lot of research attention because of their remarkable success in widespread practical applications. The implementation of DNNs on supervised learning tasks always requires a large number of training samples with accurate labels. However, in practical applications, it is too costly to label extensive data correctly, while alternating methods, such as crowdsourcing \cite{yan2014learning} and online queries \cite{schroff2011harvesting,divvala2014learning}, inexpensively obtain data, but unavoidably yield noisy labels. It is well known that training with noisy labels will degenerate the generalization performance of DNNs, which usually have the high capacity to memorize noisy labels \cite{zhang2016understanding,arpit2017closer}.

During recent years, numerous methods have been proposed to train DNNs robustly against noisy labels. Several methods focus on estimating the noise transition pattern and modifying the loss function accordingly, e.g., forward or backward correction \cite{patrini2017making}, S-model \cite{goldberger2017training}. However, it is a challenge to estimate the noise transition pattern accurately. An alternative approach is to correct labels using the predictions of DNNs, e.g., Bootstrap \cite{reed2015training}, Joint Optimization \cite{tanaka2018joint}, and D2L \cite{ma2018dimensionality}, but all of them are vulnerable to overfitting. To improve the robustness, Joint Optimization introduces regularization terms using the prior knowledge of how actual classes distribute over all training data, which is usually unavailable in practical applications.

Intuitively, we can handle noisy labels by selectively training DNNs on the clean proportion of samples \cite{han2018co,chen2019understanding} or more generally, on weighted samples \cite{ren2018learning,jiang2018mentornet}. The weight can be a real value in the interval $[0,1]$, or simply drawn from the discrete set $\{0,1\}$, where weight $1$ is assigned to examples which are believed to be clean.  Recently, several methods have been proposed based on this idea, which assign weights to training samples and minimize the weighted training loss. For example, Decoupling \cite{malach2017decoupling} trains two networks on samples for which the predictions from two networks are different. Reweight \cite{ren2018learning} assumes a clean validation set is given and assigns continuous weights to samples based on the directions of their gradients. MentorNet \cite{jiang2018mentornet} utilizes a clean validation set to pre-train a teacher network, which provides a sample weighting scheme to train a student network. When it has no access to the clean validation set, MentorNet has to assign weights according to a predefined criterion such as \textit{small-loss criterion}: treating samples with small training loss as clean ones. Co-teaching \cite{han2018co} also selects samples based on the small-loss criterion. The novelty of Co-teaching is that two networks are trained simultaneously and each network selects small-loss samples from the mini-batches and uses them to train another network.

The overfitting to noisy labels can be successfully released by the above methods, where the network assigns small weights to noisy labels, and hence the generalization accuracy can be improved. However, the practical implementations of the above methods are hindered by the lack of training samples since noisy samples are likely to be discarded from the training set. Although a sample $(x,y)$ may be noisy with $y$ being the observed label, the input $x$ is always valuable.

In this paper, we propose a novel manifold regularizer \cite{niyogi2013manifold} for DNNs, named as Paired Softmax Divergence Regularization (PSDR), to make full use of all training samples regardless of whether the observed labels are clean. 
Our regularizer is motivated from the assumption widely adopted by the Laplace Eigenmaps (LE)  \cite{belkin2003laplacian} and the Gaussian Process (GP) \cite{williams2006gaussian}, both of which assume that neighboring inputs should have similar outputs and then constrain the functions that should be learned on the training set. To design a manifold regularizer suitable for DNNs, we first clarify three questions as follows.
\begin{itemize}
	\item How to find its neighboring samples given a specific input?
	\item How to evaluate the similarity between the outputs of neighboring samples?
	\item How to implement the regularizer efficiently in DNNs to defend noisy labels?
\end{itemize}
Thus, our contributions can be summarized by answering the above questions. Firstly, since searching neighboring inputs in the whole training set is quite expensive, we propose to effectively generate neighboring samples of any specific input using the augmentation technique, which is a widely used trick to improve the generalization performance of DNNs.  Secondly, we can adopt Kullback-Leibler (KL) divergence  to evaluate the similarity between softmax outputs of paired samples generated by data augmentation, because softmax outputs naturally belong to a probability space. Finally, as a flexible regularization loss, PSDR can be easily implemented on any type of DNNs. Experiments verify that our PSDR impressively improves state-of-the-art results by a significant margin. For example, on manually corrupted CIFAR-10 with $40\%$ wrong labels, our performance is nearly as good as training on the clean CIFAR-10.

%For example, the samples from the same class are always 'close' to each other in some sense, which inspires the derivations of many unsupervised learning methods, including clustering methods and manifold learning methods. Intuitively, considering the distribution of all inputs $x$ should be beneficial to the generalization performance of DNNs. 

To systematically evaluate our proposed method, we train DNNs on noisy labels generated by manually corrupting the original ones in benchmark datasets, including CIFAR-10 and CIFAR-100 \cite{krizhevsky2009learning}, which have been widely used in the literature \cite{patrini2017making,han2018co,jiang2018mentornet,ma2018dimensionality} for evaluation of DNNs in presence of noisy labels. In the experiments, we leverage PSDR to upgrade three representative baseline methods: F-correction \cite{patrini2017making}, Decoupling \cite{malach2017decoupling}, and Co-teaching \cite{han2018co}. Empirical results consistently show that our PSDR  enables dramatically higher test accuracy, which outperforms extensive state-of-the-art methods \cite{patrini2017making,malach2017decoupling,han2018co,jiang2018mentornet,ma2018dimensionality} by quite large margins.

\section{Method}
\label{Sec_Method}
\subsection{Preliminaries}
\label{Sub_Sec_Preliminaries}
For a $c$-class classification, we collect a dataset $\mathcal{D}=\{x_t,y_t\}_{t=1}^n$, where $x_t$ is the $t$-th sample with its observed label as $y_t\in[c]:=\{1,\ldots,c\}$. As discussed previously, the observed label $y$ may be noisy. Let $\hat{y}$ denote the latent true label, then we can describe the corruption process of the set $\mathcal{D}$ by introducing a noise transition matrix $T\in\mathbb{R}^{c\times c}$, where $T_{ij}=P(y=j|\hat{y}=i)$ denotes the probability of labeling an $i$-th class example as $j$. Let $f(x;\theta)$ denote a neural network parameterized by $\theta$, which predicts a probability distribution over all classes for any input $x$. The softmax activation function is implemented on the output layer of the network to ensure $\sum_{i=1}^{c}f_i(x;\theta)=1$, where $f_i$ denotes the predicted probability for $i$-th class. The aim is to robustly train $f$ on the noisy dataset $\mathcal{D}$ by optimizing $\theta$, so that $f$ is able to predict correctly on any testing example $x$. The loss function we use is the widely adopted categorical cross entropy loss, which is denoted as $\mathcal{L}_{cce}(f(x;\theta),y)$.

\subsection{Generating neighboring samples using data augmentation}

%Without data augmentation, the theorem on finite sample expressiveness \cite{zhang2016understanding} indicates that DNNs can always achieve 0 training
%error on any finite number of training samples. \cite{zhang2016understanding} demonstrate that DNNs can fit the noisy, even random, labels contained in the training set, but the generalization error is large.

Data augmentation is a widely used technique, which can ease the overfitting problem and improve the generalization performance of DNNs. In image classification, a prevalent and effective practice for augmenting image data is to perform geometric augmentations, such as random cropping and horizontal random flipping \cite{Krizhevsky2012ImageNet,he2016deep,Perez2017The}. At each training epoch, for any training example $(x_t,y_t)$, a new input $x_t^{\prime}$ is randomly generated from $x_t$, and the loss for this example is $\mathcal{L}_{cce}(f(x_t^{\prime};\theta),y_t)$, where $\mathcal{L}_{cce}$ denotes the categorical cross entropy loss, and $\theta$ is the network parameter.

Data augmentation does ease the overfitting problem \cite{arpit2017closer}, but it does not really deal with noisy labels, and the DNNs will unavoidably memorize some wrong samples. Therefore, several specific methods of dealing with noisy labels have been proposed, among which an effective approach is learning to assign weights to training samples and minimizing the weighted training loss
\begin{equation}
\label{Eq_weight}
\mathcal{L}_{supervised}(x,y;\theta)=\sum_{t}\omega_t\mathcal{L}_{cce}(f(x_t^{\prime};\theta),y_t),
\end{equation}
where $\omega_t$ is the weight assigned to the $t$-th sample, and $x_t^{\prime}$ is the random sample generated from $x_t$ by data augmentation. The weight is usually computed in real-time during training \cite{malach2017decoupling,han2018co,jiang2018mentornet,ren2018learning}. Another representative existing method proposes to correct the categorical cross entropy loss using the noise transition matrix $T$ \cite{patrini2017making}, which can be learned from the noisy dataset, but it is a challenge to estimate $T$ accurately. For example, in F-correction \cite{patrini2017making}, the loss is
\begin{equation}
\label{Eq_correction}
\mathcal{L}_{supervised}(x,y;\theta)=\sum_{t}\mathcal{L}_{cce}(T\cdot f(x_t^{\prime};\theta),y_t).
\end{equation}
To achieve a high test accuracy, most existing methods of defending noisy labels \cite{patrini2017making,tanaka2018joint,jiang2018mentornet,ma2018dimensionality} implement data augmentation in the experiments by default.

\subsection{Paired Softmax Divergence Regularization}
Compared with normal training, recent state-of-the-art methods achieve impressive robustness when the training set contains noisy labels. However, a remaining issue is the waste of samples - existing methods do not make full use of useful information contained in the distribution of $x$. Even if a sample $(x,y)$ is noisy with $y$ being the wrong label, the input $x$ always makes sense, whose clustering contains some class-dependent information. For example, the images of the same class are `close' to each other in some sense. Intuitively, learning the information contained in the distribution of $x$ should be beneficial to the generalization performance of DNNs.

To make full use of all samples, we propose to add a manifold regularizer, which explicitly encourages the network to predict similarly on neighboring samples. Generally speaking, samples of the same class are nearby, but since the observed labels are noisy, samples with the same label are not guaranteed to be from the same latent true class. An alternative approach is searching neighboring samples in the whole training set for any specific input \cite{belkin2003laplacian}, but it is quite expensive. Fortunately, for any training example, data augmentation can generate many different neighboring images, which belong to the same class. Therefore, we explicitly penalize the difference between predictions from paired samples generated by data augmentation. Since the output of the network is a probability distribution, we use KL divergence as the penalty. Our proposed PSDR is
\begin{equation}
\label{Eq_unsupervised}
\mathcal{L}_{PSDR}(x;\theta)=\sum_{t}KL(f(x_t^{\prime};\theta)\parallel f(x_t^{\prime\prime};\theta)),
\end{equation}
where $x_t^{\prime}$ and $x_t^{\prime\prime}$ are two examples randomly generated from $x_t$ using data augmentation, and the summation is taken over the training set $\mathcal{D}=\{x_t,y_t\}_{t=1}^n$.

PSDR is a flexible regularization loss, which can be easily incorporated in existing training methods by combining $\mathcal{L}_{PSDR}$ with the supervised loss $\mathcal{L}_{supervised}$. In general, we minimize a combined loss
\begin{equation}
\label{Eq_add}
\mathcal{L}(x,y;\theta)=\mathcal{L}_{supervised}+\alpha\cdot\mathcal{L}_{PSDR}.
\end{equation}
The specific expression of $\mathcal{L}_{supervised}$ depends on the given training method, which can be simply a normal training procedure, an existing robustly training strategy expressed as Eq.~(\ref{Eq_weight}), Eq.~(\ref{Eq_correction}), or any other reasonable formulation. In this way, we can leverage PSDR to upgrade any existing methods of defending noisy labels. As an example, here we show how to improve Co-teaching \cite{han2018co} using PSDR in Algorithm~\ref{Alg1}, which is named as Co-teaching+PSDR. In Co-teaching, samples that have small training loss are selected out for training, which is equivalent to assign weights $1$ to small-loss samples and $0$ otherwise. Hence, the loss function is given by Eq.~(\ref{Eq_add}) with the $\mathcal{L}_{supervised}$ expressed as Eq.~(\ref{Eq_weight}).

\begin{algorithm}
	\caption{Co-teaching+PSDR} 
	\label{Alg1}
	\textbf{INPUT:} epoch $E_k$ and $E_{max}$, learning rate $\eta$, number of steps $N$, maximum discard ratio $\tau$, training set $\mathcal{D}$%=\{x_t,y_t\}_{t=1}^n
	\begin{algorithmic}[1]
		\STATE Initialize two networks $f_1(x;\theta_1)$ and $f_2(x;\theta_2)$
		\FOR {$e=1,2,\cdots,E_{max}$}
		\STATE Shuffle the training set $\mathcal{D}$
			\FOR {$i=1,2,\cdots,N$}
			\STATE Draw mini-batch $\bar{\mathcal{D}}$ from $\mathcal{D}$, where each sample $x_t$ in $\bar{\mathcal{D}}$ is augmented to have $x_t^{\prime}$ and $x_t^{\prime\prime}$
			\STATE Sample $R(e)\%$ small-loss samples with $f_1$: $\quad\bar{\mathcal{D}}_{f_1}=\{R(e)\%\,\arg\min_{\bar{\mathcal{D}}}\mathcal{L}_{cce}(f_1(x_t^{\prime}),y_t)\}$
			\STATE Sample $R(e)\%$ small-loss samples with $f_2$: $\quad\bar{\mathcal{D}}_{f_2}=\{R(e)\%\,\arg\min_{\bar{\mathcal{D}}}\mathcal{L}_{cce}(f_2(x_t^{\prime}),y_t)\}$
			\STATE $\theta_1=\theta_1-\eta\bigtriangledown_{\theta_1}(\sum\limits_{\bar{\mathcal{D}}_{f_2}}\mathcal{L}_{cce}(f_1(x_t^{\prime}),y_t)+\alpha\sum\limits_{\bar{\mathcal{D}}}KL(f_1(x_t^{\prime})\parallel f_1(x_t^{\prime\prime})))$
			\STATE $\theta_2=\theta_2-\eta\bigtriangledown_{\theta_2}(\sum\limits_{\bar{\mathcal{D}}_{f_1}}\mathcal{L}_{cce}(f_2(x_t^{\prime}),y_t)+\alpha\sum\limits_{\bar{\mathcal{D}}}KL(f_2(x_t^{\prime})\parallel f_2(x_t^{\prime\prime})))$
			%\STATE Update $\theta_1$ by decending the loss:\\ %\centerline{$\,\sum\limits_{\bar{\mathcal{D}}_{f_2}}\mathcal{L}_{cce}(f_1(x_t^{\prime}),y_t)+\alpha\sum\limits_{\bar{\mathcal{D}}}KL(f_1(x_t^{\prime})\parallel f_1(x_t^{\prime\prime}))$}
			%\STATE Update $\theta_2$ by backpropagating the loss:\\ %\centerline{$\,\sum\limits_{\bar{\mathcal{D}}_{f_1}}\mathcal{L}_{cce}(f_2(x_t^{\prime}),y_t)+\alpha\sum\limits_{\bar{\mathcal{D}}}KL(f_2(x_t^{\prime})\parallel f_2(x_t^{\prime\prime}))$}
			\ENDFOR
			\STATE Update $R(e)=1-\min\{\frac{e}{E_k}\tau,\tau\}$
		\ENDFOR
	\end{algorithmic}
	\textbf{OUTPUT:} The trained models, $f_1(x;\theta_1)$ and $f_2(x;\theta_2)$	
\end{algorithm}

\section{Experiments}
\label{Sec_Experiments}
In this section, we empirically verify the effectiveness of our proposed PSDR. Notably, compared with extensive state-of-the-art methods \cite{patrini2017making,malach2017decoupling,han2018co,jiang2018mentornet,ma2018dimensionality}, PSDR enables the best test accuracy when the training set contains noisy labels. Specifically, we demonstrate that PSDR significantly improves three representative baseline methods: F-correction \cite{patrini2017making}, Decoupling \cite{malach2017decoupling}, and Co-teaching \cite{han2018co}, and the improved ones also outperform other baseline methods.

\subsection{Experimental setup}

\begin{figure}[t]
	%\vskip 0.2in
	\begin{center}
		\centerline{\includegraphics[width=0.5\columnwidth]{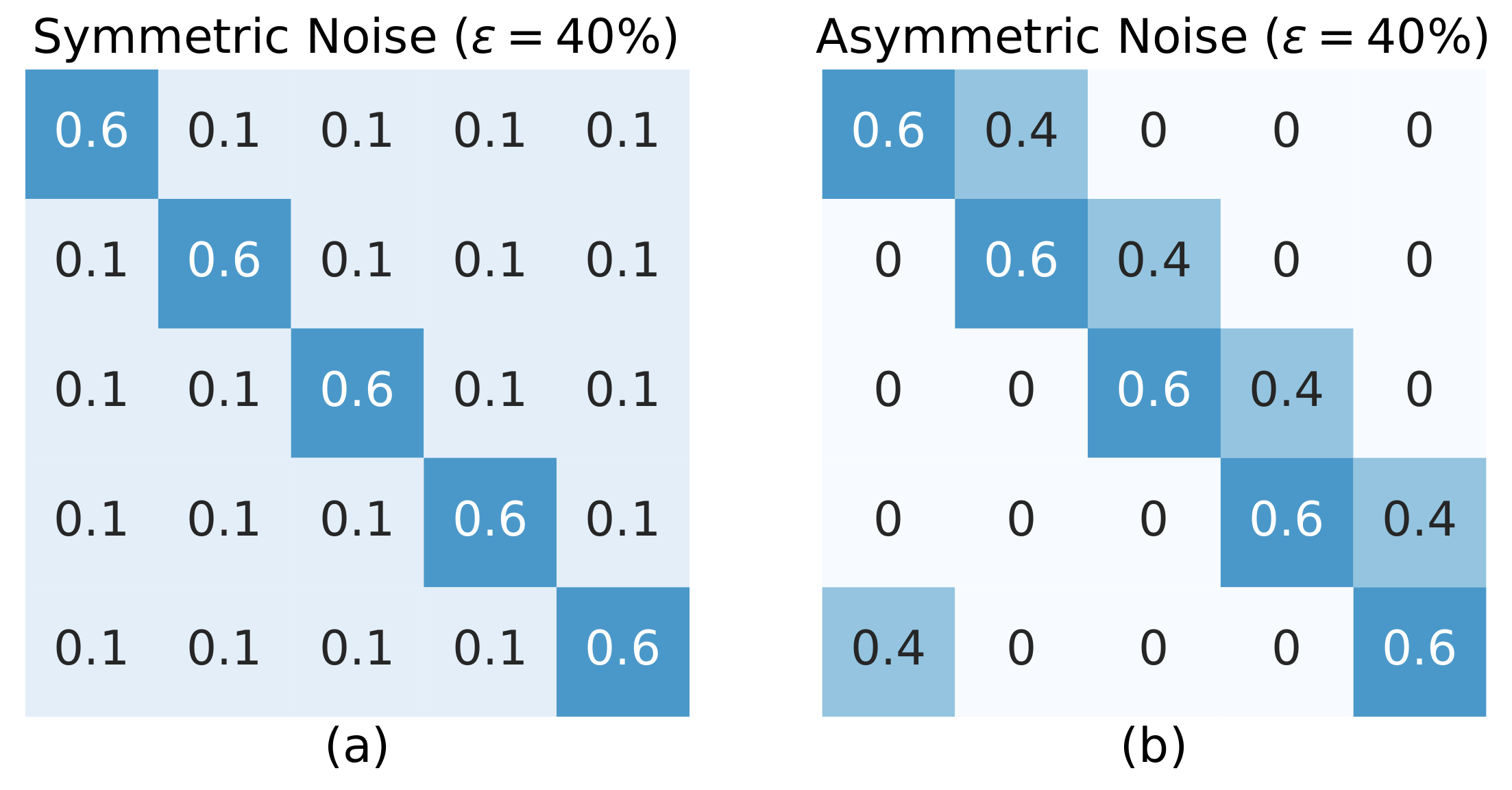}}
		\vskip -0.1in
		\caption{Examples of noise transition matrix $T$ for symmetric and asymmetric noise (taking 5 classes and noise ratio $\varepsilon=40\%$ as an example).}
		\label{Fig_T}
	\end{center}
	%\vskip -0.2in
\end{figure}

\subsubsection{Datasets and noise structures}
Our method is verified on the benchmark datasets CIFAR-10 and CIFAR-100 \cite{krizhevsky2009learning}, which are widely used in the literature \cite{patrini2017making,han2018co,jiang2018mentornet,ma2018dimensionality} for evaluation of DNNs in presence of noisy labels. Since the labels in CIFAR-10 and CIFAR-100 are taken as ground truth, the noisy labels are generated by randomly flipping the original ones. Following previous literature \cite{ren2018learning,han2018co,jiang2018mentornet,ma2018dimensionality}, we test on two representative types of noise: symmetric noise and asymmetric noise. As illustrated in Fig.~\ref{Fig_T}, in the symmetric case, all labels can flip uniformly to any other classes, while in the asymmetric case, labels in a class can flip to a single class. The noise ratio $\varepsilon$ denotes the proportion of wrong labels. Note that asymmetric noise with ratio $50\%$ is trivial, hence in the experiments, we test symmetric noise with ratio $20\%$, $50\%$, and asymmetric noise with ratio $40\%$.

\subsubsection{Compared methods} We investigate the following baselines. ($1$) \textit{F-correction} \cite{patrini2017making}. It first trains a network to estimate $T$, then modifies the loss function accordingly. ($2$) \textit{Decoupling} \cite{malach2017decoupling}. It trains two networks on samples for which the predictions from the two networks are different. ($3$) \textit{Co-teaching} \cite{han2018co}. It maintains two networks. Each network selects samples of small training loss from the mini-batches and uses them to train another network. ($4$) \textit{MentorNet} \cite{jiang2018mentornet}. A teacher network is pre-trained, which provides a sample weighting scheme to train a student network. ($5$) \textit{Joint Optimization} \cite{tanaka2018joint}. It updates observed labels using predictions of the network. ($6$) \textit{D2L} \cite{ma2018dimensionality}. For each sample, it linearly combines the original label and the label predicted by the network. The combining weight is computed using the dimensionality of the latent feature subspace \cite{amsaleg2017vulnerability}. In the experiments, standard data augmentation is applied to all methods.

\subsubsection{Training details}
For fair comparisons, all methods are evaluated with the same setup. To ensure that the empirical results are reliable, we repeat each experiment $5$ times and report the average test accuracy. Following the official implementation of ResNet \cite{he2016deep} in Keras, we train the ResNet-32 \cite{he2016deep} for $200$ epochs with a batch size of $128$. We use the Adam optimizer \cite{kinga2015method} with an initial learning rate $10^{-3}$, which is divided by $10$ after $80$, $120$ and $160$ epochs, and further divided by $2$ after $180$ epochs. In the network, we implement $l_2$ weight decay of $10^{-4}$. We use the standard data augmentation: horizontal random flipping and $32\times32$ random cropping after padding 4 pixels around images. In our method, the combining factor $\alpha$ in Eq.~(\ref{Eq_add}) is simply set to $1$. Although tuning $\alpha$ may result in better performance, we show that $\alpha=1$ is sufficient for beating all baselines.

\subsection{Results on CIFAR-10}

\begin{figure*}[t]
	%\vskip 0.2in
	\begin{center}
		\centerline{\includegraphics[width=0.9\columnwidth]{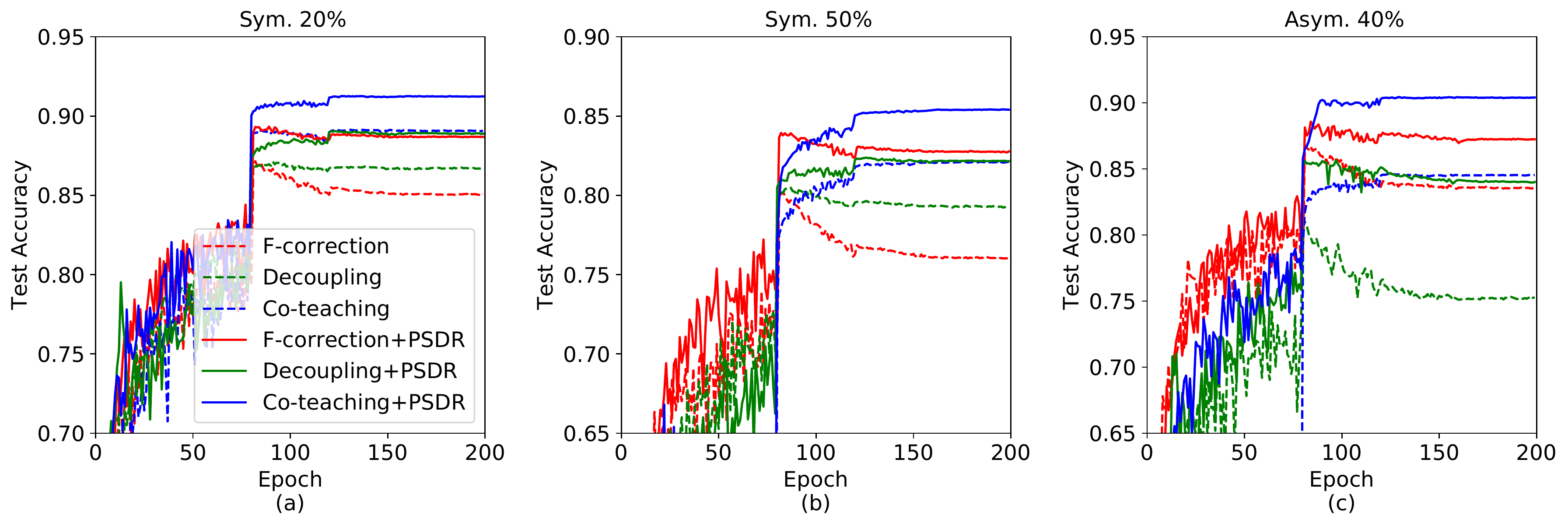}}
		%\vskip -0.15in
		\caption{Average test accuracy (5 runs) on CIFAR-10 during training under symmetric noise with ratio $20\%$, $50\%$, and asymmetric noise with ratio $40\%$. We train the ResNet-32 on manually corrupted CIFAR-10 and test on the clean test set. The sharp change of test accuracy results from learning rate change.%We plot the test accuracy after the first learning rate change at $80$ epoch.
		}
		\label{Fig_cifar10}
	\end{center}
	%\vskip -0.2in
\end{figure*}

\begin{table}[t]
	\caption{Average test accuracy ($\%$, 5 runs) on CIFAR-10 under symmetric noise with ratio $20\%$, $50\%$, and asymmetric noise with ratio $40\%$. We train the ResNet-32 on manually corrupted CIFAR-10 and test on the clean test set. The best test accuracy under each setting is marked in bold face. The clean baseline means normal training on the clean CIFAR-10 without corruption.}
	\label{Tab_cifar10}
	\centering
	\begin{tabular}{|c|c|c|c|}
		\hline
		\multicolumn{1}{|c|}{\multirow{2}{*}{Method}}
		& \multicolumn{2}{c|}{Symmetric}    & Asymmetric\\ \cline{2-4} 
		& $20\%$  & $50\%$  & $40\%$  \\ \hline
		MentorNet         & $88.36\pm0.46$         & $77.10\pm0.44$         & $77.33\pm0.79$          \\
		Joint Optimization& $85.30\pm0.35$         & $79.84\pm1.18$         & $84.34\pm1.37$          \\
		D2L               & $86.12\pm0.43$         & $67.39\pm13.62$         & $85.57\pm1.21$          \\ \hline
		F-correction      & $85.08\pm0.43$         & $76.02\pm0.19$         & $83.55\pm2.15$          \\
		F-correction+PSDR & $88.68\pm0.28$ 		& $82.77\pm0.52$		  & $87.25\pm1.78$          \\ \hline
		Decoupling        & $86.72\pm0.32$         & $79.31\pm0.62$         & $75.27\pm0.83$          \\
		Decoupling+PSDR   & $88.89\pm0.63$			& $82.16\pm1.05$		  & $84.01\pm0.69$          \\ \hline
		Co-teaching       & $89.05\pm0.32$         & $82.12\pm0.59$         & $84.55\pm2.81$          \\
		Co-teaching+PSDR  & $\textbf{91.24}\pm0.19$ & $\textbf{85.40}\pm0.66$ & $\textbf{90.41}\pm0.37$ \\ \hline
		Clean baseline 	  & \multicolumn{3}{c|}{$91.55$}        				   \\ \hline
	\end{tabular}
\end{table}

\begin{figure*}[t]
	%\vskip 0.2in
	\begin{center}
		\centerline{\includegraphics[width=0.9\columnwidth]{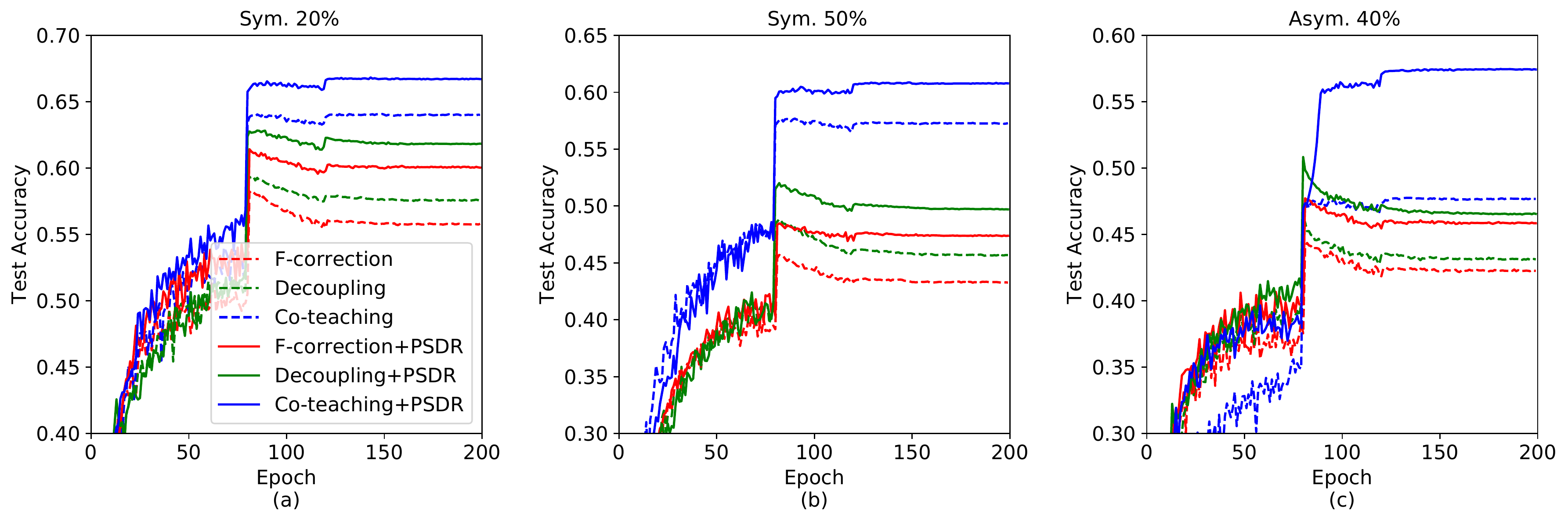}}
		%\vskip -0.15in
		\caption{Average test accuracy (5 runs) on CIFAR-100 during training under symmetric noise with ratio $20\%$, $50\%$, and asymmetric noise with ratio $40\%$. We train the ResNet-32 on manually corrupted CIFAR-100 and test on the clean test set. The sharp change of test accuracy results from learning rate change.%We plot the test accuracy after the first learning rate change at $80$ epoch.
		}
		\label{Fig_cifar100}
	\end{center}
	%\vskip -0.2in
\end{figure*}

\begin{table}[t]
	\caption{Average test accuracy ($\%$, 5 runs) on CIFAR-100 under symmetric noise with ratio $20\%$, $50\%$, and asymmetric noise with ratio $40\%$. We train the ResNet-32 on manually corrupted CIFAR-100 and test on the clean test set. The best test accuracy under each setting is marked in bold face. The clean baseline means normal training on the clean CIFAR-100 without corruption.}
	\label{Tab_cifar100}
	\centering
	\begin{tabular}{|c|c|c|c|}
		\hline
		\multicolumn{1}{|c|}{\multirow{2}{*}{Method}}
		& \multicolumn{2}{c|}{Symmetric}    & Asymmetric\\ \cline{2-4} 
		& $20\%$  & $50\%$  & $40\%$  \\ \hline
		MentorNet         & $62.96\pm0.44$         & $46.38\pm0.35$         & $42.37\pm0.47$          \\
		Joint Optimization& $28.36\pm1.71$         & $25.41\pm1.06$         & $23.18\pm1.97$          \\
		D2L               & $12.47\pm4.19$         & $\ \ 5.62\pm5.43$         & $14.08\pm5.81$       \\ \hline
		F-correction      & $55.80\pm0.54$         & $43.27\pm0.72$         & $42.25\pm0.73$          \\
		F-correction+PSDR & $60.04\pm0.49$			& $47.38\pm1.14$		  & $45.85\pm0.89$          \\ \hline
		Decoupling        & $57.55\pm0.47$         & $45.68\pm0.43$         & $43.12\pm0.42$          \\
		Decoupling+PSDR   & $61.84\pm0.26$			& $49.70\pm0.88$		  & $46.55\pm0.36$          \\ \hline
		Co-teaching       & $64.02\pm0.26$         & $57.27\pm0.36$         & $47.67\pm1.24$          \\
		Co-teaching+PSDR  & $\textbf{66.71}\pm0.44$ & $\textbf{60.78}\pm0.52$ & $\textbf{57.43}\pm0.21$ \\ \hline
		Clean baseline 	  & \multicolumn{3}{c|}{$67.06$}       				       \\ \hline
	\end{tabular}
	\vskip -0.1in
\end{table}

In Table~\ref{Tab_cifar10}, we report the test accuracy on the clean test set. We show the results for five recent successful baselines, and as an example, we use PSDR to further upgrade three of them. As shown in the table, PSDR consistently enables existing methods to achieve much higher test accuracy. Specifically, Co-teaching+PSDR achieves the best generalization performance under all noise settings, as marked in bold face. For comparison, we also normally train the same ResNet-32 on the original CIFAR-10 without any corruption and report the test accuracy in Table~\ref{Tab_cifar10}, which is $91.55\%$. Notably, the performance of Co-teaching+PSDR is very close to the clean baseline under symmetric noise with ratio $20\%$ and asymmetric noise with ratio $40\%$. Even under symmetric noise with ratio $50\%$, where half of the labels are wrong, Co-teaching+PSDR achieves an impressive test accuracy of $85.4\%$.

In Fig.~\ref{Fig_cifar10}, we show the test accuracy during training. Specifically, we are interested in how PSDR affects the convergence of DNNs, so we plot the accuracy after the first learning rate change at the $80$th epoch. As we can see, without PSDR, the networks suffer from severe overfitting to noisy labels, which is indicated by the decrease of test accuracy at the later stage of training (e.g., the dashed green curve in Fig.~\ref{Fig_cifar10} (c)). As shown in the figure, the overfitting is eased by PSDR. For all investigated noise settings and baseline methods, PSDR consistently improves the convergence of DNNs.

\subsection{Results on CIFAR-100}
In Table~\ref{Tab_cifar100}, we report the test accuracy for CIFAR-100. The observations here are consist with those for CIFAR-10. PSDR consistently improves the generalization performance of investigated baseline methods, and the improved ones also outperform other state-of-the-art methods. As marked in bold face, the best result is achieved by Co-teaching+PSDR. For further comparison, we also normally train the same ResNet-32 on the original CIFAR-100 without any corruption and report the test accuracy in Table~\ref{Tab_cifar100}, which is $67.06\%$. Impressively, the test accuracy of Co-teaching+PSDR is very close to the clean baseline under symmetric noise with ratio $20\%$. For other noise settings, Co-teaching+PSDR also significantly outperforms other methods.

To investigate how PSDR affects the convergence of DNNs, we show the test accuracy during training in Fig.~\ref{Fig_cifar100}. As can be seen, PSDR reduces the decrease of test accuracy at the later stage of training, which implies it ease the overfitting to noisy labels. In all experiments, PSDR consistently improves the convergence of DNNs compared with baseline methods.

\section{Discussion}
\label{Sec_Discussion}

\subsection{How does smoothness benefit defending noisy labels?}
\begin{figure}[t]
	%\vskip 0.2in
	\begin{center}
		\centerline{\includegraphics[width=0.9\columnwidth]{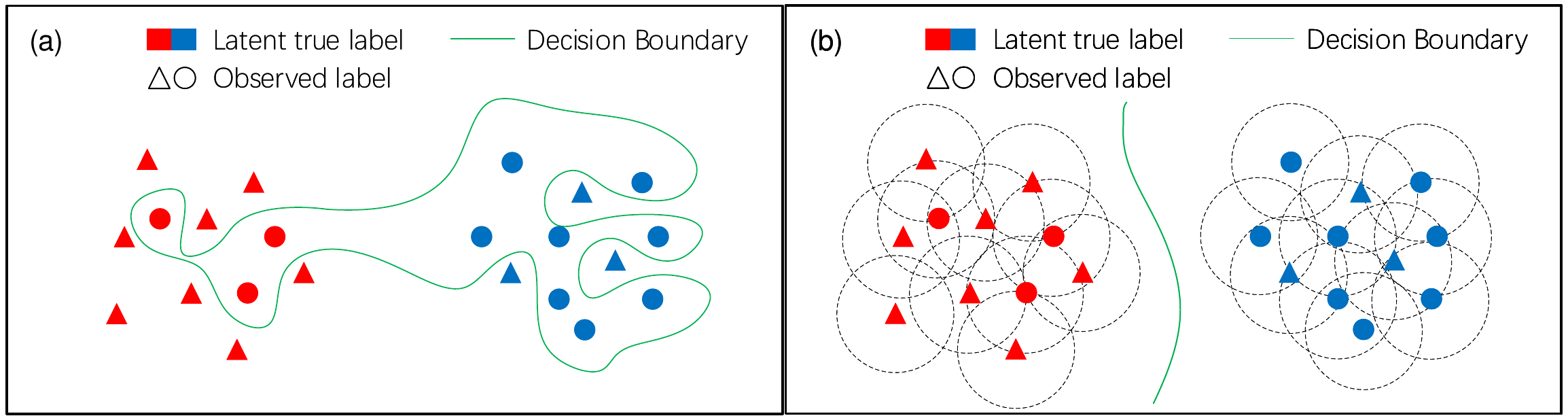}}
		\caption{Illustration of decision boundary when trained with noisy labels. The the color \textit{red} and \textit{blue} indicate the latent true label, and the shape \textit{triangle} and \textit{circle} indicate the observed label. We use the \textit{dashed circle} to indicate the area of data augmentation for each sample.
		}
		\label{Fig_illustration}
	\end{center}
	\vskip -0.2in
\end{figure}

To simplify our presentation, we define the smoothness as predicting similarly on samples of the same class. This assumption has been widely used in manifold learning \cite{belkin2003laplacian}, but is never adopted to improve the generalization performance of DNNs trained on noisy datasets. As demonstrated empirically in this paper, we find that smoothness on samples of the same class is beneficial to robustness against noisy labels. Without loss of generality, we consider the non-trivial case such that $\forall i$, $T_{ii}$ is the largest among $\{T_{ij},j=1,\cdots,c\}$, which means for samples with actual class $i$, the number of correct labels $i$ is the largest among all labels $\{1,\cdots,c\}$. In this case, if we can enforce smoothness on all samples with actual class $i$, then the network will be very likely to correctly predict class $i$ for these samples since the number of samples with label $i$ is the largest. However, in our setting, the observed labels are noisy, samples with the same label are not guaranteed to be from the same latent true class, so we can't implement this idea to enforce smoothness according to the observed labels. Moreover, searching neighboring samples of a given input as \cite{belkin2003laplacian} is quite expensive. Therefore, we utilize data augmentation, which can generate many different neighboring images, and the generated images are guaranteed to be in the same class.

The idea can be further illustrated in Fig.~\ref{Fig_illustration}, where we use different colors to indicate the true labels and different shapes to indicate the observed noisy labels. Without any regularization, the theorem on finite sample expressiveness \cite{zhang2016understanding} implies that DNNs can always achieve 0 training error on any finite number of training samples, as illustrated in Fig.~\ref{Fig_illustration} (a). In this case, the model overfits to the noisy labels, which usually degenerates the generalization performance. On the other hand, with data augmentation, we can generate many neighboring samples of the same class. For illustration, in Fig.~\ref{Fig_illustration} (b), we use the dashed circle to indicate the area of data augmentation for each input sample. If we can enforce smoothness on the augmented samples, then it is impossible for the DNNs to learn a complex decision boundary that separates nearby samples into multiple classification regions. Since the samples with correct labels are the majority in each class, the network is very likely to correctly predict the actual class.

\subsection{PSDR enforces smoothness explicitly}
Data augmentation is a widely used technique which eases the overfitting problem and improves the generalization performance of DNNs \cite{Krizhevsky2012ImageNet,he2016deep,Perez2017The}. Traditionally, at each epoch, an augmented sample is randomly generated for each sample in the original training set, and the network is trained on the augmented ones. In this way, for any original sample, various random samples are generated during the whole training process. Since the augmented samples have the same label as the original example, directly training on them enforces smoothness implicitly around original ones. 

\begin{figure}[t]
	%\vskip 0.2in
	\begin{center}
		\centerline{\includegraphics[width=0.3\columnwidth]{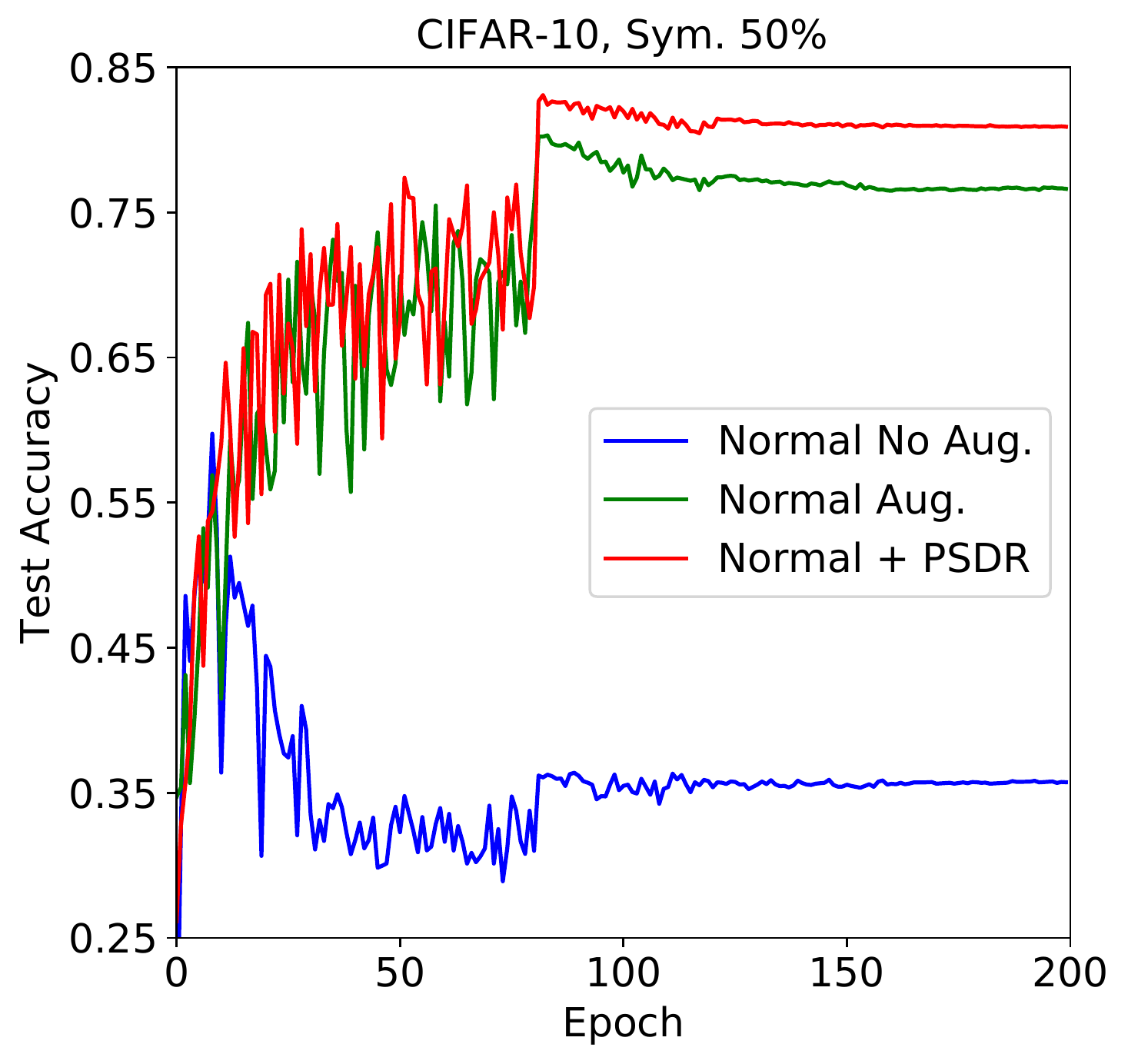}}
		\vskip -0.1in
		\caption{Normal training with (i) data augmentation turned off (Normal No Aug.), (ii) data augmentation turned on by default (Normal Aug.), and (iii) PSDR (Normal + PSDR). As an example, we train the ResNet-32 on manually corrupted CIFAR-10 that has symmetric noise with ratio $50\%$.
		%Data augmentation is beneficial to reducing overfitting, and PSDR further improves the robustness against noisy labels. The sharp change of test accuracy results from learning rate change.
		}
		\label{Fig_discussion}
	\end{center}
	\vskip -0.3in
\end{figure}

In Fig.~\ref{Fig_discussion}, we show the test accuracy of DNNs w.r.t. the number of training epochs. In the experiments, we normally train the DNNs without any specific techniques of dealing with noisy labels. As an example, we train on manually corrupted CIFAR-10 that has symmetric noise with ratio $50\%$. As shown in the blue curve, without data augmentation, the test accuracy first reaches a high value, then decreases quickly, which implies the network eventually overfit to the noisy labels. This phenomenon is consistent with the finding in \cite{arpit2017closer}, which states that DNNs learn simple patterns first, then memorize noisy labels. The green curve in Fig.~\ref{Fig_discussion} shows the test accuracy of the network directly trained on augmented samples. As we can see, data augmentation improves the test accuracy significantly, although overfitting still occurs at the later stage of training.

PSDR further improves the robustness to noisy labels by enforcing smoothness explicitly, as shown in Fig.~\ref{Fig_discussion}. Recall that in PSDR, the KL divergence between softmax outputs of paired samples generated by data augmentation is directly added to the training loss as a manifold regularization. In this way, PSDR enforces smoothness explicitly. Moreover, PSDR is an unsupervised regularization term, which enables all samples to make sense during training, regardless of whether their labels are clean. PSDR overcomes the problem of existing methods: many baseline methods do not make full use of all samples, and some of them simply discard the likely noisy samples, which results in lack of training samples, and hence reduces the generalization performance. Fig.~\ref{Fig_discussion} demonstrates that with PSDR, even if the network is normally trained without any other specific methods of defending noisy labels, we still can achieve a test accuracy of around $80\%$ on the manually corrupted CIFAR-10 containing $50\%$ wrong labels.

\section{Conclusion}
\label{Sec_Conclusion}
In this paper, we present a simple but effective regularization called PSDR, which significantly improves the robustness of DNNs trained with noisy labels. Our method is motivated by the fact that many existing methods do not make full use of all training samples. By encouraging the predictions to be similar for paired samples generated using data augmentation, PSDR enables all training samples from the original training set to make sense in training, regardless of whether their labels are clean. We conduct comprehensive experiments on benchmark datasets under different noise types and noise ratios. Empirical results verify that our PSDR consistently improves existing state-of-the-art methods by a significant margin.

%\subsubsection*{Acknowledgments}

%\clearpage
\bibliography{neurips_2019}

\begin{thebibliography}{}

\bibitem[Amsaleg et~al., 2017]{amsaleg2017vulnerability}
Amsaleg, L., Bailey, J., Barbe, D., Erfani, S., Houle, M.~E., Nguyen, V., and
  Radovanovi{\'c}, M. (2017).
\newblock The vulnerability of learning to adversarial perturbation increases
  with intrinsic dimensionality.
\newblock {\em Information Forensics and Security (WIFS)}, pages 1--6.

\bibitem[Arpit et~al., 2017]{arpit2017closer}
Arpit, D., Jastrz{\k{e}}bski, S., Ballas, N., Krueger, D., Bengio, E., Kanwal,
  M.~S., Maharaj, T., Fischer, A., Courville, A., Bengio, Y., et~al. (2017).
\newblock A closer look at memorization in deep networks.
\newblock In {\em ICML}.

\bibitem[Belkin and Niyogi, 2003]{belkin2003laplacian}
Belkin, M. and Niyogi, P. (2003).
\newblock Laplacian eigenmaps for dimensionality reduction and data
  representation.
\newblock {\em Neural computation}, 15(6):1373--1396.

\bibitem[Chen et~al., 2019]{chen2019understanding}
Chen, P., Liao, B.~B., Chen, G., and Zhang, S. (2019).
\newblock Understanding and utilizing deep neural networks trained with noisy
  labels.
\newblock In {\em ICML}.

\bibitem[Divvala et~al., 2014]{divvala2014learning}
Divvala, S.~K., Farhadi, A., and Guestrin, C. (2014).
\newblock Learning everything about anything: Webly-supervised visual concept
  learning.
\newblock In {\em CVPR}.

\bibitem[Goldberger and Ben-Reuven, 2017]{goldberger2017training}
Goldberger, J. and Ben-Reuven, E. (2017).
\newblock Training deep neural-networks using a noise adaptation layer.
\newblock In {\em ICLR}.

\bibitem[Han et~al., 2018]{han2018co}
Han, B., Yao, Q., Yu, X., Niu, G., Xu, M., Hu, W., Tsang, I., and Sugiyama, M.
  (2018).
\newblock Co-teaching: robust training deep neural networks with extremely
  noisy labels.
\newblock In {\em NeurIPS}.

\bibitem[He et~al., 2016]{he2016deep}
He, K., Zhang, X., Ren, S., and Sun, J. (2016).
\newblock Deep residual learning for image recognition.
\newblock In {\em CVPR}.

\bibitem[Jiang et~al., 2018]{jiang2018mentornet}
Jiang, L., Zhou, Z., Leung, T., Li, L.-J., and Fei-Fei, L. (2018).
\newblock Mentornet: Learning data-driven curriculum for very deep neural
  networks on corrupted labels.
\newblock In {\em ICML}.

\bibitem[Kinga and Adam, 2015]{kinga2015method}
Kinga, D. and Adam, J.~B. (2015).
\newblock A method for stochastic optimization.
\newblock In {\em ICLR}.

\bibitem[Krizhevsky and Hinton, 2009]{krizhevsky2009learning}
Krizhevsky, A. and Hinton, G. (2009).
\newblock Learning multiple layers of features from tiny images.
\newblock Technical report, Citeseer.

\bibitem[Krizhevsky et~al., 2012]{Krizhevsky2012ImageNet}
Krizhevsky, A., Sutskever, I., and Hinton, G.~E. (2012).
\newblock Imagenet classification with deep convolutional neural networks.
\newblock In {\em NeurIPS}.

\bibitem[Ma et~al., 2018]{ma2018dimensionality}
Ma, X., Wang, Y., Houle, M.~E., Zhou, S., Erfani, S.~M., Xia, S.-T.,
  Wijewickrema, S., and Bailey, J. (2018).
\newblock Dimensionality-driven learning with noisy labels.
\newblock In {\em ICML}.

\bibitem[Malach and Shalev-Shwartz, 2017]{malach2017decoupling}
Malach, E. and Shalev-Shwartz, S. (2017).
\newblock Decoupling" when to update" from" how to update".
\newblock In {\em NeurIPS}.

\bibitem[Niyogi, 2013]{niyogi2013manifold}
Niyogi, P. (2013).
\newblock Manifold regularization and semi-supervised learning: Some
  theoretical analyses.
\newblock {\em The Journal of Machine Learning Research}, 14(1):1229--1250.

\bibitem[Patrini et~al., 2017]{patrini2017making}
Patrini, G., Rozza, A., Menon, A.~K., Nock, R., and Qu, L. (2017).
\newblock Making deep neural networks robust to label noise: A loss correction
  approach.
\newblock In {\em CVPR}.

\bibitem[Perez and Wang, 2017]{Perez2017The}
Perez, L. and Wang, J. (2017).
\newblock The effectiveness of data augmentation in image classification using
  deep learning.
\newblock In {\em NeurIPS}.

\bibitem[Reed et~al., 2015]{reed2015training}
Reed, S., Lee, H., Anguelov, D., Szegedy, C., Erhan, D., and Rabinovich, A.
  (2015).
\newblock Training deep neural networks on noisy labels with bootstrapping.
\newblock In {\em ICLR}.

\bibitem[Ren et~al., 2018]{ren2018learning}
Ren, M., Zeng, W., Yang, B., and Urtasun, R. (2018).
\newblock Learning to reweight examples for robust deep learning.
\newblock In {\em ICML}.

\bibitem[Schroff et~al., 2011]{schroff2011harvesting}
Schroff, F., Criminisi, A., and Zisserman, A. (2011).
\newblock Harvesting image databases from the web.
\newblock 33(4):754--766.

\bibitem[Tanaka et~al., 2018]{tanaka2018joint}
Tanaka, D., Ikami, D., Yamasaki, T., and Aizawa, K. (2018).
\newblock Joint optimization framework for learning with noisy labels.
\newblock In {\em CVPR}.

\bibitem[Williams and Rasmussen, 2006]{williams2006gaussian}
Williams, C.~K. and Rasmussen, C.~E. (2006).
\newblock {\em Gaussian processes for machine learning}, volume~2.
\newblock MIT Press Cambridge, MA.

\bibitem[Yan et~al., 2014]{yan2014learning}
Yan, Y., Rosales, R., Fung, G., Subramanian, R., and Dy, J. (2014).
\newblock Learning from multiple annotators with varying expertise.
\newblock {\em Machine learning}, 95(3):291--327.

\bibitem[Zhang et~al., 2017]{zhang2016understanding}
Zhang, C., Bengio, S., Hardt, M., Recht, B., and Vinyals, O. (2017).
\newblock Understanding deep learning requires rethinking generalization.
\newblock In {\em ICLR}.

\end{thebibliography}
\bibliographystyle{apalike}

\medskip

\small

\end{document}